\documentclass[runningheads]{llncs}
\usepackage{graphicx}
\usepackage{multirow}
\usepackage{float}

\begin{document}
\title{UNER: Universal Named-Entity Recognition Framework}
%
%
\author{Diego Alves\inst{1}\and
Tin Kuculo\inst{2}\and
Gabriel Amaral\inst{3}\and
Gaurish Thakkar\inst{1}\and
Marko Tadić\inst{1}
}
\authorrunning{D. Alves et al.}

%
\institute{Faculty of Humanities and Social Sciences, University of Zagreb, Zagreb 10000, Croatia
\email{\{dfvalio,marko.tadic\}@ffzg.hr},
\email{gthakkar@m.ffzg.hr}
\and
L3S Research Center, Leibniz University Hannover, Hannover, Germany
\email{kuculo@l3s.de} \and
King's College London, London, United Kingdom\\
\email{gabriel.amaral@kcl.ac.uk}}

\maketitle              
\begin{abstract}
We introduce the Universal Named-Entity Recognition (UNER) framework, a 4-level classification hierarchy, and the methodology that is being adopted to create the first multilingual UNER corpus: the SETimes parallel corpus annotated for named-entities. First, the English SETimes corpus will be annotated using existing tools and knowledge bases. After evaluating the resulting annotations through crowdsourcing campaigns, they will be propagated automatically to other languages within the SETimes corpora. Finally, as an extrinsic evaluation, the UNER multilingual dataset will be used to train and test available NER tools. As part of future research directions, we aim to increase the number of languages in the UNER corpus and to investigate possible ways of integrating UNER with available knowledge graphs to improve named-entity recognition. 

\end{abstract}
%
%
\section{Introduction}

In the span of little more than a year, with pre-trained language models becoming ubiquitous in the field of Natural Language Processing (NLP), a wide range of tasks have received renewed interest; not the least of which is information extraction. Information extraction (IE) is the task of automatically extracting structured information from natural language texts. A common sub-task of information extraction includes Named Entity Recognition and Classification (NERC) \footnote[1]{Copyright © 2020 for this paper by its authors. Use permitted under Creative Commons License Attribution 4.0 International (CC BY 4.0).}.

NERC corpora usually respond to the specific needs of local projects in terms of the complexity of the annotation hierarchy and its format. Table \ref{tab1} shows some of the various NERC annotation schemes that have been proposed in previous research efforts. 

\begin{table}
\caption{Description of existing NE annotation schemes in terms of hierarchy complexity.}\label{tab1}
\begin{tabular}{|l|r|r|}
\hline
\textbf{Source} &  \textbf{Number of Levels} & \textbf{Nodes per level}\\
\hline
MUC-7\cite{chinchor-robinson-1998-appendix} (English) & 2  & 3/8\\
CoNLL 2003 \cite{sang2003introduction} (English and German) & 1 & 4\\
spaCy \cite{spacy2} (based on OntoNote5, English) & 1 & 18\\
Czech Named Entity Corpus 2.0 \cite{vsevvcikova2007named} (Czech) &  2 & 8/46\\
NKJP corpus\cite{przepiorkowski2011national} (Polish) & 2 & 6/8\\
Second Harem \cite{freitas2010second} (Portuguese) & 3 & 10/36/21\\
Sekine \cite{sekinewebsite} (English v.7.1.0) & 4 & 3/28/87/125\\
\hline
\end{tabular}
\end{table}

With the aim of creating a universal multilingual annotation NE scheme, we follow 
previous work (Table \ref{tab1}) to propose the Universal Named Entity Recognition (UNER) framework, a four-level NE classification hierarchy.

We intend to use the proposed framework to annotate the multilingual SETimes parallel corpora\footnote[2]{http://nlp.ffzg.hr/resources/corpora/setimes/} (described in \cite{tyers2010south}) and create a multilingual named entity recognition dataset. 

This paper is organised as follows: In Section 2 we describe our approach to defining the UNER hierarchy; Section 3 describes our strategy for the creation of the first multilingual open-source UNER corpora; and in Section 4 we provide conclusions and possible future directions for research, including applications of UNER corpora and increasing UNER multilingualism.

\section{UNER Tagging Framework Definition}

The UNER hierarchy is built upon the NER hierarchy proposed by Sekine \cite{sekinewebsite}, which presents the highest conceptual complexity between the compared NER schemes (Table \ref{tab1}). Maintaining its main structure, UNER is  organised through one root node ("TOP"), and three first-level leaf nodes ("Name", "Timex TOP" and "Numex") which correspond to MUC-7\cite{chinchor-robinson-1998-appendix} main categories. Analysing each of the child nodes, we make the following changes to compose the UNER hierarchy:

\begin{itemize}
    \item In Sekine's hierarchy, "Person" is a child node of "Name" and has no ramifications. In UNER, "Name Other" and "God" have become child nodes of "Person", their names changed to "Other" and "Entity". In addition to that, "Person" gets new child nodes: "Name", "Profession" and "Fictional".
    
    \item Concerning the leaf node "Location" (inside "Name"), we have introduced a new child node "Fictional" and have removed "Phone Number" which was added as a child node of "Numex".
    
    \item Inside the "Product" node's ramifications, we have moved "ID Number" to the "Numex" node, suppressed "Character" and "Title" as these nodes were replaced by the "Profession" and "Fictional" nodes respectively, inside the parent node "Person". We have also added a child leaf node "Brand" inside the nodes "Clothing", "Drugs", "Food", "Vehicles" and "Weapon". Brands that do not correspond to these categories will be annotated as "Product Other", a child node of "Product". 
    
    \item Inside "Event", a child node of "Name", we have introduced a new child node "Personal" concerning personal facts such as births and weddings. We have also renamed the node "Incident" as "Historical Event" and have added a child node named "Other" inside it. In Sekine's hierarchy, historical events (e.g. French Revolution) were classified as "Event Other" inside "Event". 
    
    \item Concerning the leaf node "Timex TOP" which corresponds to time expressions,  we have added "Holiday" as a child node inside "Timex". We have also introduced a child node named "Timex Relative" with the same ramifications as the node "Timex". The idea is to differentiate between absolute time expressions such as "April 1st, 2003" and relative expressions as in "last August".
\end{itemize}

We have focused only on the nodes inside Sekine's hierarchy; the attributes proposed by the authors to increase the knowledge (for example: "Place of Origin" as an attribute of the node "River") were not taken into account in this version of the UNER framework. 


As previously explained, UNER follows Sekine's NER hierarchy. It is composed of a root node "TOP" and three child nodes: "Name", "Timex TOP" and "Numex", each of which contain several ramifications. The number of nodes of each UNER level is described in Table \ref{tab3}.

Due to the size of the UNER hierarchy scheme, we have decided to present it fully in digital form\footnote[3]{https://tinyurl.com/sb3u9ve}.

\begin{table}
\caption{Description of the number of nodes per level inside UNER hierarchy.}\label{tab3}
\centering
\begin{tabular}{|l|r|}
\hline
\textbf{Level} & \textbf{Number of nodes}\\
\hline
0 (root) & 1\\
1 & 3 \\
2 & 29\\
3 &  95\\
4 & 129\\
\hline
\end{tabular}
\end{table}

\subsection{Format}
Available open-source NER tools can be broadly classified into 3 main types based on the format they support. These are BIO (Begin-Inside-Outside), inline XML-tags ($<$PERSON$>$ George Clooney $<$/PERSON$>$), or Index-based spans (text:'George Clooney', start:0, end:14, label:'Person'). Since we want the dataset to support existing training frameworks, we performed a study on existing available NER tools and their corresponding support for various input formats. This is summarised in Table \ref{tab2}. The input column is the support of the tool for taking in data for training a new model and output denotes the final output format when new text is passed through the tagging process. Cells marked as "-" refer to tools that do not support training and only have a prediction functionality. In the deployment phase of UNER, we intend to use BIO and Index-based span format to represent the NEs, but all formats are convertible into each other with simple scripts, which is important to guarantee its universality.

\begin{table}
\caption{Existing NER tools and their support for various formats.}\label{tab2}
\centering
  \begin{tabular}{|c|c|c|c|c|c|c|}
    \hline
    \multirow{2}{*}{\textbf{Tool}} &
      \multicolumn{2}{c}{\textbf{BIO}} &
      \multicolumn{2}{c}{\textbf{XML}} &
      \multicolumn{2}{c|}{\textbf{Index-based}} \\
      
    & Input & Output & Input & Output& Input & Output \\
    \hline
    Polyglot \cite{chen2014building} & - & Yes & - & -&- &-\\
    \hline
NER BERT \cite{larionov2019semantic} & Yes& Yes & No&No &No &No\\
    \hline
Stanford NER  \cite{manning-EtAl:2014:P14-5}  & Yes & Yes& No& Yes &No &Yes\\
    \hline
ESNLTK \cite{ORASMAA16.332} & No&Yes & No& No& Yes&Yes\\
     \hline
Finnish-tagtools \cite{Sam2018}& -&No & -&Yes &- &No\\
     \hline
Spacy \cite{spacy2}& Yes&Yes & No&No &Yes &Yes\\
     \hline
Poldeepner \cite{marcinczuk2018recognition}& Yes & No& No& No& No&Yes\\
     \hline
BILSTM-CRF-CHAR \cite{de2018lener} & Yes & Yes&No &No &No &No\\
     \hline
Stagger \cite{ostling2013stagger}& -&No & No&Yes &- &No\\
     \hline
Swener \cite{kokkinakis2014hfst}& -&No &- &Yes & -&No\\
     \hline
Flair \cite{akbik2018coling} & Yes&Yes & No& No& Yes&Yes\\
    \hline
  \end{tabular}
\end{table}

\section{Application methodology}

The methodology that will be adopted to create the 10 multilingual UNER annotated corpora is schematized in Figure 1. 

\begin{figure}
\includegraphics[width=\textwidth]{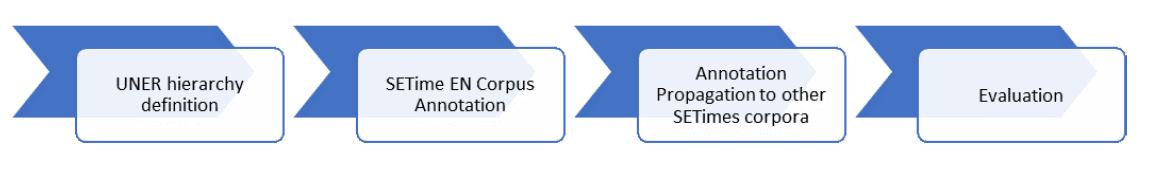}
\caption{Steps at the application of the UNER framework methodology to a multilingual parallel corpus.} \label{fig1}
\end{figure}

Using the defined UNER hierarchy, we will tag the SETimes English corpus by combining automatic annotation methods. Following this, we will use crowdsourcing campaigns to evaluate the quality of the annotated data, and to provide insight on possible improvements to our annotation methods. These annotations will then be propagated to the other languages in the SETimes corpora, and the generated data will be used to train new NER models that will be evaluated in terms of precision, recall, and F-measure.

The SETimes parallel corpus (CC-BY-SA license) is based on the contents published on the SETimes.com news portal, which published “news and views from Southeast Europe” and it covers ten languages: Bulgarian, Bosnian, Greek, English, Croatian, Macedonian, Romanian, Albanian, Serbian and Turkish. In this way we can quickly build UNER systems for many languages and check the universality of the proposed scheme. The English corpus is composed of 4,248,417 words and 205,910 sentences\footnote[4]{The corrected version of SETimes corpus where diacritics and encoding system has been corrected and is available from nlp.ffzg.hr. We will consider this version in our study.}.

The process of annotating the English SETimes corpus consists of a hybrid computation workflow. 

We employed state-of-the-art NERC tools to pre-annotate the corpus. We use spaCy \cite{spacy2} (3 models each with 18 tags) and Flair \cite{akbik2018coling} (1 model with 4 classes and 2 models 18 classes) to tag the corpus. This step was performed by sorting all the previously mentioned models according to their reported recall, establishing a priority order. We then iterated through each model according to its priority and annotated the corpora. For each iteration, we checked all the tokens that were left untagged by the previous models and used the current one to complete the annotations. The aim of this step is to maximise the overall number of annotated entities, thus, increasing the overall recall. Using this method we have identified 631,068 entity occurrences inside the English SETimes corpus which correspond to 89,600 different character strings.  

Then, using SPARQWrapper, we will correct the tags by retrieving information from DBpedia \cite{auer2007dbpedia}, Yago \cite{suchanek2007yago} and Wikidata \cite{vrandevcic2014wikidata}. For that, we will define precise one-to-one correspondences between the relevant classes of these databases and UNER nodes. 

The quality of the annotation will be evaluated through crowdsourcing tasks and, lastly, we will propagate the corrected annotations from the English corpus to those in other languages. This final process will be done by employing existing label propagation algorithms and models, such as graph propagation methods \cite{tamura2012bilingual} and machine-translation models \cite{jain2019entity}.
 
Once the annotations are propagated from the English corpus to the other 9 parallel SETimes datasets, we aim to conduct an extrinsic evaluation of the generated data by using it to train and test NERC deep learning models (for example Stanford NLP NERC module based on Conditional Random Fields). 

For each language, a deep analysis of the entity distribution inside the annotated corpus will be conducted to determine possible biases that may influence evaluation, also taking into consideration missing examples for some categories. 

\section{Conclusions and Future Directions}
We have presented the UNER framework, a universal multilingual hierarchy building on Sekine's NER hierarchy \cite{sekinewebsite}, to be used as a universal framework for NERC annotations. We have also designed a hybrid computation workflow for the creation, correction and possible evaluation of an annotated parallel multilingual corpora using UNER as an annotation framework and the SETimes corpus as the basis. The realization of this workflow and the delivery of the annotated corpora are intended as future work. Similarly, we intend to implement automatic methods for the extraction of events, according to the ACE 2005 event annotation guidelines \cite{ace05Event}. 

We also intend to apply the UNER framework to new languages and evaluate its value as a training corpus for cross-lingual and multilingual named entity and event extraction. Finally, we aim to employ the resulting trained models using UNER corpora in combination with multilingual knowledge graphs, such as Wikidata \cite{vrandevcic2014wikidata}, for the creation of NERC sub-models for multilingual and multicultural environments.

\section{Acknowledgements}
The work presented in this paper has received funding from the European Union’s Horizon 2020 research and innovation programme under the Marie Skłodowska-Curie grant agreement no. 812997 and under the name CLEOPATRA (Cross-lingual Event-centric Open Analytics Research Academy). 

%
%
%
\bibliographystyle{splncs04}
\bibliography{mybibliography}
%
\end{document}